\documentclass[review,5p,twocolumn]{elsarticle}

\usepackage{hyperref}

\journal{Artificial Intelligence in Medicine}

\usepackage{times}  
\usepackage{url}  
\usepackage{graphicx}  

\usepackage{amsmath}
\usepackage{amsfonts}
\usepackage{tabularx}
\usepackage{multirow}
\usepackage{makecell}

\usepackage[labelfont=bf,justification=raggedright,singlelinecheck=false]{caption}
\usepackage[flushleft]{threeparttable}

\usepackage[figuresright]{rotating}
\usepackage{booktabs}

\biboptions{sort&compress}

\allowdisplaybreaks[4] 

\begin{document}

\begin{frontmatter}

\title{Classifying medical relations in clinical text via convolutional neural networks}

\author[a]{Bin He}
\address[a]{Research Center of Language Technology, Harbin Institute of Technology, Harbin, China}
\ead{hebin\_hit@hotmail.com}
\author[a]{Yi Guan\corref{cor}}
\cortext[cor]{Corresponding author}
\ead{guanyi@hit.edu.cn}
\author[b]{Rui Dai}
\address[b]{Department of Mathematics, Harbin Institute of Technology, Harbin, China}
\ead{13B912003@hit.edu.cn}

\begin{abstract}
Deep learning research on relation classification has achieved solid performance in the general domain. This study proposes a convolutional neural network (CNN) architecture with a multi-pooling operation for medical relation classification on clinical records and explores a loss function with a category-level constraint matrix. Experiments using the 2010 i2b2/VA relation corpus demonstrate these models, which do not depend on any external features, outperform previous single-model methods and our best model is competitive with the existing ensemble-based method.
\end{abstract}


\end{frontmatter}


\section{Introduction}

Relation classification, a natural language processing (NLP) task identifying the relation between two entities in a sentence, is an important technique used in many subsequent NLP applications such as question answering and knowledge base completion. This task has been widely studied in the general domain due to the large number of accessible datasets such as the SemEval-2010 task 8 dataset \cite{Hendrickx2010}, which aims to classify the relation between two nominals in the same sentence.

In the clinical domain, Informatics for Integrating Biology and the Bedside (i2b2) released an annotated relation corpus on clinical records, attracting considerable attention \cite{Uzuner2011}. Identifying relations in clinical records is more challenging than relations in the general domain because one sentence from clinical records may contain more than two medical concepts and concepts may be comprised of several words. For example, the sentence ``at that time , she also had \emph{cat scratch fever} and she had \emph{resection} of \emph{an abscess in the left lower extremity}" contains three concepts, two of which contain more than two words. The annotated information given in the 2010 i2b2/VA relation corpus thus differs from that in the SemEval-2010 task 8 dataset. In the former, the category to which a concept pair belongs is given, and the classification objective is to identify the subcategory, also known as the \emph{relation type}.

Deep neural networks have become a research trend in recent years due to powerful learning ability features without manual feature engineering. Various neural architectures have been proposed for classifying relations in general \cite{Socher2012,Zeng2014,DBLP:conf/acl/SantosXZ15,Wang2016}, biomedical \cite{Liu2016,Zhao2016,Quan2016,Asada2017,sahu2017drug,Zheng2017} and clinical text \cite{Sahu2016Relation,Raj2017}. Conventional convolutional neural network (CNN) models use max-pooling operations to extract the most significant feature in a convolutional filter; however, information regarding feature positioning relative to the concepts cannot be captured. Responding to this issue, \citet{ChenXLZ015} designed a dynamic multi-pooling method to extract features from each part of a sentence in the argument classification task. A chunk-based max-pooling algorithm, proposed by \cite{Zhang2015}, splits each sentence into a fixed number of segments to retain more semantics from the sentence for the statistical machine translation model. The position of features relative to concepts is vital for medical relation classification on clinical records. Based on the above studies, this study proposed a CNN-based method (without any external features) for recognizing medical concept relations in clinical records. Its contributions are as follows:
\begin{itemize}
  \item A multi-pooling operation was introduced into the proposed CNN architecture, which aims to capture the position information of local features relative to the concept pair.
  \item A novel loss function with a category-level constraint matrix was explored.
  \item The proposed models achieved improved performance compared to previous single-model methods, and the best model is competitive with the ensemble-based method for classifying relations between medical concepts.
\end{itemize}

\section{Corpus and data preprocessing}
\label{sec:corpus}

The relation corpus\footnote{The relation dataset is available at \url{https://www.i2b2.org/NLP/Relations/}.} used in this study was released in the 2010 i2b2/VA challenge, and is comprised of 426 discharge summaries. Of these, 170 were used for training, and the remaining 256 for testing\footnote{This follows the official data split in the 2010 i2b2/VA challenge.}. This dataset contains three types of concepts (\emph{medical problem}, \emph{treatment}, and \emph{test}), and each concept pair in the same sentence was assigned one relation type. Medical concept relations in this corpus can be grouped into 3 categories: \emph{medical problem-treatment}, \emph{medical problem-test}, and \emph{medical problem-medical problem} relations. Table~\ref{tab:relationcount} describes the definitions\footnote{2010 i2b2/VA Challenge Evaluation Relation Annotation Guidelines: \url{http://www.i2b2.org/NLP/Relations/assets/Relation\%20Annotation\%20Guideline.pdf}.} and statistics of these relation types.

\begin{table}[!htb]
  \footnotesize
  \caption{\label{tab:relationcount}Relation type statistics.}
  \begin{threeparttable}
    \begin{tabularx}{\linewidth}{lXll}
      \hline
      Relation & Definition & Train & Test \\ \hline
      \multicolumn{4}{l}{\emph{Medical problem-Treatment relations}} \\
      TrIP & Treatment improves medical problem & 51 & 152 \\
      TrWP & Treatment worsens medical problem & 24 & 109 \\
      TrCP & Treatment causes medical problem & 184 & 342 \\
      TrAP & Treatment is administered for medical problem & 885 & 1732 \\
      TrNAP & Treatment is not administered because of medical problem & 62 & 112 \\
      NTrP & No relation between treatment and problem & 1702 & 2759 \\ \hline
      \multicolumn{4}{l}{\emph{Medical problem-Test relations}} \\
      TeRP & Test reveals medical problem & 993 & 2060 \\
      TeCP & Test conducted to investigate medical problem & 166 & 338 \\
      NTeP & No relation between test and problem & 993 & 1974 \\ \hline
      \multicolumn{4}{l}{\emph{Medical problem-Medical problem relations}} \\
      PIP & Medical problem indicates medical problem & 755 & 1448 \\
      NPP & No relation between two medical problems & 4418 & 8089 \\ \hline
    \end{tabularx}
    \begin{tablenotes}[para,flushleft]
      Eight positive relation types were annotated in this relation corpus, and samples of three negative relation types (starting with ``N" in this table) were extracted for model training to ensure each concept pair within a sentence could be classified into a certain relation type.
    \end{tablenotes}
  \end{threeparttable}
\end{table}

Although words within sentences were already separated by spaces, additional splits were required for some specific strings. This study employed the Natural Language Toolkit\footnote{Natural Language Toolkit: \url{http://www.nltk.org/}.} (NLTK) to tokenize sentence strings in clinical records, then realigned concept boundaries to avoid concept information errors. Tokens\footnote{Definition of \emph{token}: \url{http://nlp.stanford.edu/IR-book/html/htmledition/tokenization-1.html}.} were lowercase, and numbers were replaced by zero. 

\begin{figure}[!htb]
    \centering
    \includegraphics[width=\linewidth]{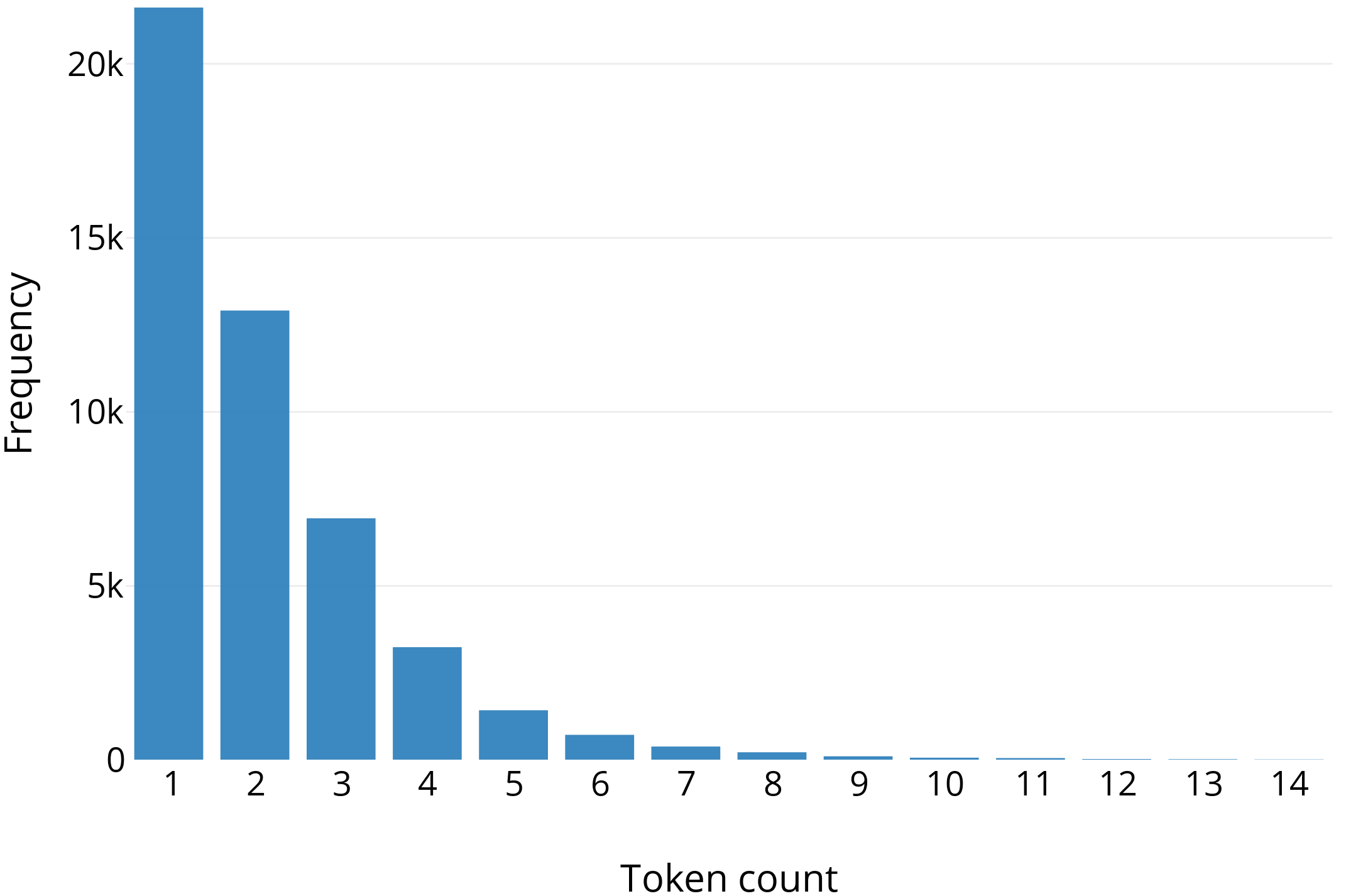}
    \caption{\label{fig:entlen}Frequency distribution of token count in medical concepts. Concept lengths appearing less than five times were filtered.}
\end{figure}

\begin{figure*}[!htb]
    \centering
    \includegraphics[width=\linewidth]{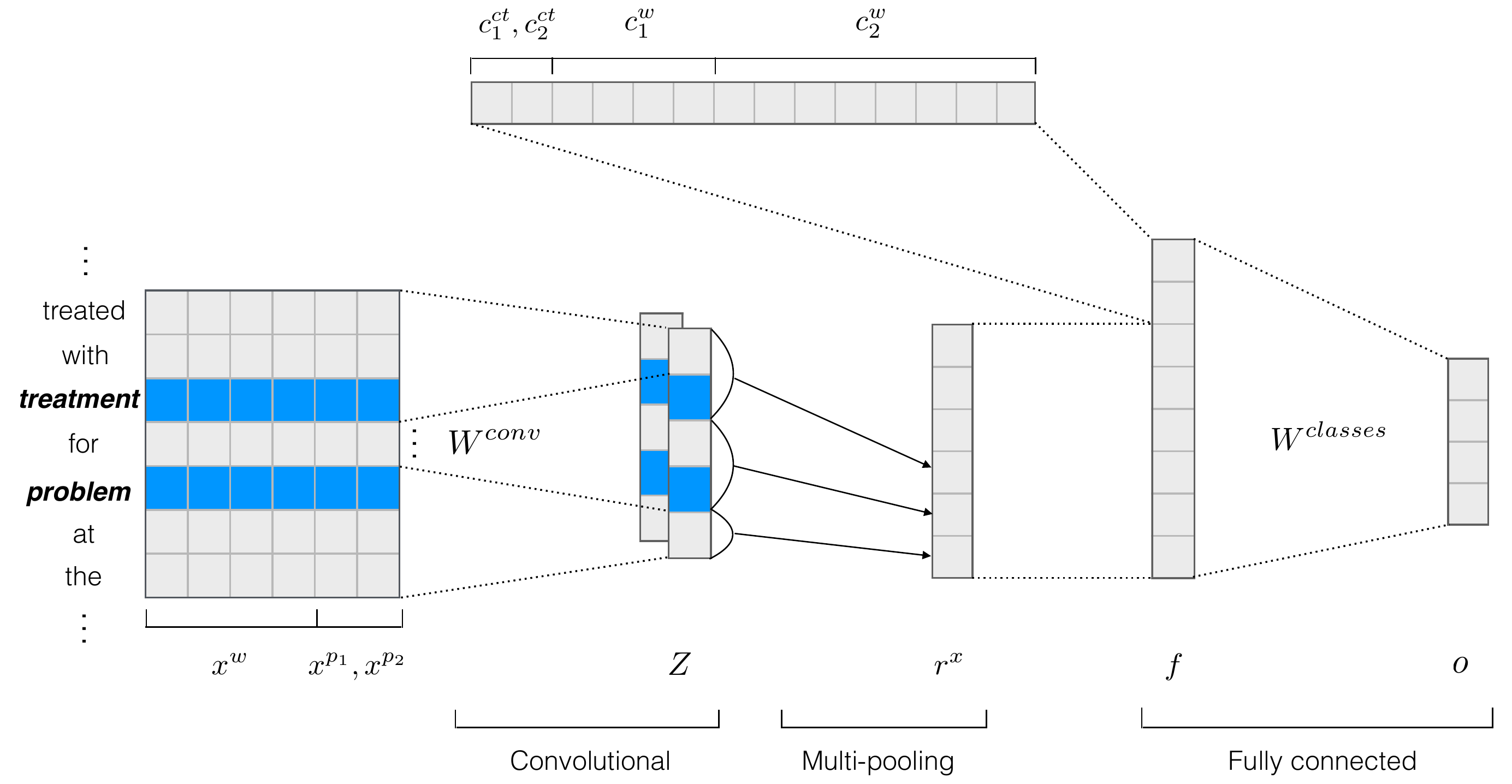}
    \caption{\label{fig:arch}Architecture of CNN-based model for medical relation classification. Concept contents in the sample sentence ``she was treated with $[\text{steroids}]_{treatment}$ for $[\text{this\ swelling}]_{medical\ problem}$ at the outside hospital , and these were continued ." are replaced by their concept types.}
\end{figure*}

Fig.~\ref{fig:entlen} lists the frequency distribution of token count in medical concepts, showing over half the concepts contained more than one token. According to these statistics, the average token count for all concepts in the corpus was 2.09 (\emph{medical problem} 2.42, \emph{treatment} 1.85, and \emph{test} 1.86, respectively). Considering concepts containing more than one token were detrimental to n-gram feature extraction in the proposed model, the concept pair contents were replaced by their concept types to make local features more apparent. The details of the replacement are shown in Fig.~\ref{fig:arch}.

\section{Methodology}

Fig.~\ref{fig:arch} describes the architecture of the proposed CNN-based model for medical relation classification on clinical records. This model learns a distributed representation for each relation sample. A feature vector is generated to represent each sentence sample $x$ containing two concepts, and final scores are calculated with relation type representations. Further details are discussed in the following subsections.

\subsection{Word representation}
\label{sec:wordrepresentation}

In this relation classification task, the following information was given for each sample sentence $x=(x_1,x_2,\dots,x_n)$: (1) concept position indexes in the sentence $c_1^{index}$ and $c_2^{index}$; (2) concept contents $c_1=(c_{11},\dots,c_{1l_1})$ and $c_2=(c_{21},\dots,c_{2l_2})$; (3) concept types $c_1^{type}$ and $c_2^{type}$; and (4) $y$ is the sample's relation type.

Previous studies on relation classification \cite{Zeng2014,DBLP:conf/acl/SantosXZ15,Wang2016,Sahu2016Relation} utilized word position features to capture information on the proximity of words to target concepts. Therefore, an word embedding matrix $W^{w}\in\mathbb{R}^{d^w\times |V^w|}$ and an word position embedding matrix $W^{wp}\in\mathbb{R}^{d^p\times |V^p|}$ were utilized in this study, where $V^w$ represented the vocabulary, $V^p$ represented the word position set, and $d^w$ and $d^p$ were pre-set embedding sizes. For each sample sentence, every word was mapped to a column vector $x^w_i\in\mathbb{R}^{d^w}$ representing the word feature. Additionally, the relative distances between the current word and concepts were mapped to the position vectors $x^{p_1}_i\in\mathbb{R}^{d^p}$ and $x^{p_2}_i\in\mathbb{R}^{d^p}$. Based on these features, each word could be represented as $x_i'=[(x^w_i)^T, (x^{p_1}_i)^T, (x^{p_2}_i)^T]^T\in\mathbb{R}^{d^x}$, where $d^x$ was the word vector size and $d^x=d^w+2d^p$.

\subsection{Convolutional multi-pooling}

Semantic representations of n-grams are valuable features in relation classification tasks, and convolution operation can capture this information by combining word representations in a fixed window. Given a sample sentence $x=(x_1,x_2,\dots,x_n)$ and a context window size $k$, the concatenation of successive words in this window size could be defined as:
\[x_{i:i+k-1}=[(x_i')^T,\dots,(x_{i+k-1}')^T]^T,\]
and $x_{i:i+k-1}\in\mathbb{R}^{d^xk}$. The representation of this sentence could be reformatted as $X=(x_{1:k},\dots,x_{n-k+1:n})$ and $X\in\mathbb{R}^{d^xk\times (n-k+1)}$. The input $X$ would then be fed into the convolutional layer to generate local features. Given $W^{conv}$ as the weight matrix of the convolutional filters and a linear bias $B_1$, a linear transformation followed by a non-linear function are calculated:
\[Z=f(W^{conv}\cdot X + B_1),\]
where $W^{conv}\in\mathbb{R}^{d^c\times d^xk}$, $B_1\in\mathbb{R}^{d^c}$, $f$ is the relu function, and the convolutional result is $Z\in\mathbb{R}^{d^c\times (n-k+1)}$.

Max-pooling operations are generally used to extract the most significant feature in a convolutional filter \cite{Zeng2014}; however, these are insufficient for relation classification on clinical records. The dataset used here contains $\sim$3.3 concepts in one sentence, making the position of features relative to the concept pair necessary for relation classification. A multi-pooling operation was introduced in the proposed method to achieve more local features in each sentence. Although word position information was included in word representations, multi-pooling strengthened the significance of the relative position information.

Given the concept position index of a concept pair described in Section \ref{sec:wordrepresentation}, the convolutional result $Z$ can be split into three parts: $Z^1=Z_{1:(c_1^{index}-1)}$, $Z^2=Z_{c_1^{index}:(c_2^{index}-1)}$, and $Z^3=Z_{c_2^{index}:(n-k+1)}$, where $Z_{p:q}=[Z_p,\dots,Z_q]\in\mathbb{R}^{d^c\times (q-p+1)}$. Max-pooling operations are then performed on each part to extract the most valuable features, defined as $r^l_j=max[Z^l]_j$, $r^l\in\mathbb{R}^{d^c}$, and $l\in\{1,2,3\}$. These three vectors can be concatenate into the single vector
\[r^x=[(r^1)^T, (r^2)^T, (r^3)^T]^T\in\mathbb{R}^{3d^c},\]
creating an informative semantic representation of the sentence.

\subsection{Concept feature representation}

As described in Section \ref{sec:corpus}, the concept types in a concept pair are given, allowing their relation category to be known directly. In response to this situation, concept type information is typically used in two ways: to train multiple independent models, or to train one model by adding the concept types as features. Both methods have distinct advantages and disadvantages: the former cannot maintain unified word representation, and each model loses some samples to update the word embedding matrix; the latter may produce misclassifications across categories.

In order to maintain unified word representation and tend to model simplicity, the latter method was selected here for model building, and two vectors were used to represent two concept types mapped from a concept type embedding matrix $W^{ct}\in\mathbb{R}^{d^{ct}\times |V^{ct}|}$. In the matrix, $V^{ct}$ represents concept type set and $d^{ct}$ represents a pre-set concept type embedding size. This concept type feature representation was formalized as $x^{ct}=[(c_1^{ct})^T, (c_2^{ct})^T]^T\in\mathbb{R}^{2d^{ct}}$. Concept content, in addition to the n-gram and concept type features described above, is also necessary for the relation classification model. Word embeddings of the concept contents were added to supplement concept feature representation, which can be formalized as $cf^x=[(x^{ct})^T, (c_1^w)^T, (c_2^w)^T]^T\in\mathbb{R}^{d^{cf}}$, where $c_i^w=[(c_{i1}^w)^T,\dots,(c_{il_i}^w)^T]^T$ is the concept content representation, $i\in\{1,2\}$, $c_{ij}^w$ is the word representation of the $j$th word in the $i$th concept, and $d^{cf}=2d^{ct}+d^w\times(l_1+l_2)$.

\subsection{Class embeddings and scoring}

The n-gram feature representation and concept feature representation were concatenated into the single vector $rc = [(r^x)^T, (cf^x)^T]^T$, and the confidence of each relation type with a class embedding matrix $W^{classes}\in\mathbb{R}^{m\times (3d^c+d^{cf})}$ was computed as
\[s = W^{classes}\cdot rc,\]
where each row vector $W^{classes}_l$ can be viewed as the representation of relation type $l$ and $m$ equals the number of relation types.

\paragraph{Training with logsoftmax} After obtaining relation type scores, a softmax operation was applied to obtain the probability of each relation type:
\[
p(y|x,\theta)=\frac{e^{s_y}}{\sum_{l\in\mathcal{Y}}{e^{s_l}}},
\]
where $s_y$ is the score for the relation type $y$, $\mathcal{Y}$ is the relation type set, and $\theta=(W^w, W^{wp}, W^{conv}, B_1, W^{ct}, W^{classes})$. Based on this probability, the loss function could be defined as
\begin{align*}
  \mathcal{L} = &-\log p(y|x, \theta) + \beta(||W^w||^2+||W^{wp}||^2+||W^{conv}||^2\\
  &+||W^{ct}||^2+||W^{classes}||^2),
\end{align*}
and $\beta$ was the $L_2$ regularization parameter.

\paragraph{Category-level constraint} Training one model to cover all categories may cause cross-category misclassifications. It would also be inappropriate to regard samples in other categories as negative samples. Therefore, a loss function with a category-level constraint matrix was proposed:
\begin{align*}
  \mathcal{L}_C = &\log (\sum_{l\in\mathcal{Y}}{C_{l'l'}^x \cdot e^{s_l}}) - s_y + \beta(||C^x\cdot W^{classes}||^2\\
  &+||W^w||^2+||W^{wp}||^2+||W^{conv}||^2+||W^{ct}||^2),
\end{align*}
\[
C_{ij}^x = \begin{cases}
1, & \text{if } i=j \text{ and } i \in \text{Category}_x;\\
0, & \text{otherwise}.
\end{cases}
\]
Here, $C^x$ represents the constraint matrix of relation type indexes, $l'$ represents the index number of relation type $l$, and $\text{Category}_x$ represents the relation type index set for the category that sample $x$ belongs to. After using this loss function during the training of sample $x$, only the class vectors $W^{classes}_l$($l' \in \text{Category}_x$) will be updated, and the other class vectors remain unchanged. This avoids treating samples in other categories as negative.

\section{Experiments}

\subsection{Experimental setup}
\label{sec:es}

\paragraph{Evaluation metric} As shown in Table~\ref{tab:relationcount}, there are eight positive relation types and three negative relation types. Precision, recall, and F1-measure were used to evaluate the performance of each positive relation type. Simultaneously, as stipulated in the official evaluation metric \cite{Uzuner2011}, model performance was defined based on the micro-averaged F1 score across all positive relation types. 

\paragraph{Parameter settings} Initial word representations were trained using the word2vec tool \cite{Mikolov2013} and de-identified notes from the MIMIC-III database \cite{Johnson2016}. The other matrices in the proposed method were randomly initialized by normalized initialization \cite{DBLP:journals/jmlr/GlorotB10}. The word embedding size was set to 50 and the concept type embedding size to 5, equal to those in \cite{Sahu2016Relation}. The dropout technique \cite{Srivastava2014} was used in the concatenated representation $rc$ to avoid overfitting, and this value was set to 0.5. One fifth of the training set was randomly selected as the development set during experiments, and the model hyperparameters were tuned using a grid search: word position embedding size (5, 10, 20, 30); convolutional filter size (100, 200, 300, 400); learning rate (0.01, 0.025, 0.05, 0.075, 0.1); $L_2$ regularization parameter (0.00005, 0.0001, 0.0005, 0.001). The selected hyperparameter values were 10, 200, 0.075, and 0.0005, respectively.

\subsection{Experimental results}

Three method comparisons were designed: (1) CNN-Max, the CNN-based model using max-pooling in the convolutional layer; (2) CNN-Multi, the CNN-based model using multi-pooling in the convolutional layer; and (3) CNN-Multi-C, where the CNN-Multi model was trained with category-level constraint.

\subsubsection{Filter window sizes and word embedding initializations}

The efficacy of different filter window sizes and word embedding initializations was investigated using the CNN-Multi model. For each filter window size, model performance was evaluated under two word embedding initializations: (1) pre-trained, where word embeddings are initialized by pre-trained word embeddings as described in Section~\ref{sec:es}; and (2) randomly initialized, where word embeddings are randomly initialized by normalized initialization \cite{DBLP:journals/jmlr/GlorotB10}. Table~\ref{tab:wordemb} shows the system performance by measures of precision, recall, and F1 score.

\begin{table}[!htb]
  \centering
  \footnotesize
  \caption{\label{tab:wordemb}CNN-Multi model performance using various convolutional window sizes and different word embedding initializations.}
    \begin{tabularx}{\linewidth}{lXXXXXX}
      \hline & \multicolumn{3}{l}{Pre-trained} & \multicolumn{3}{l}{Randomly initialized} \\ \cmidrule(lr){2-4}\cmidrule(lr){5-7}
      Window size & P & R & F1 & P & R & F1 \\ \hline
      {[}3] & 73.7 & 64.1 & 68.5 & 73.2 & 65.8 & 69.3 \\
      {[}4] & 72.2 & 63.6 & 67.6 & 73.1 & 66.7 & \bf 69.7 \\
      {[}5] & 72.7 & 62.7 & 67.3 & 74.5 & 62.3 & 67.9 \\
      {[}3,4] & 73.9 & 61.6 & 67.2 & 71.0 & 67.5 & 69.2 \\
      {[}3,5] & 73.2 & 62.1 & 67.2 & 71.2 & 67.1 & 69.1 \\
      {[}4,5] & 71.7 & 65.6 & 68.5 & 75.3 & 61.7 & 67.8 \\
      {[}3,4,5] & 70.7 & 67.0 & 68.8 & 70.8 & 67.7 & 69.2 \\
      \hline
    \end{tabularx}
\end{table}

Pre-trained word embeddings demonstrated lower F1 scores in most window sizes. The highest F1 score was achieved using a window size 4 and randomly initialized word embeddings. Therefore, all proposed models were trained using a filter window size 4 and randomly initialized word embeddings.

\begin{figure*}[!htb]
    \centering
    \includegraphics[width=\linewidth]{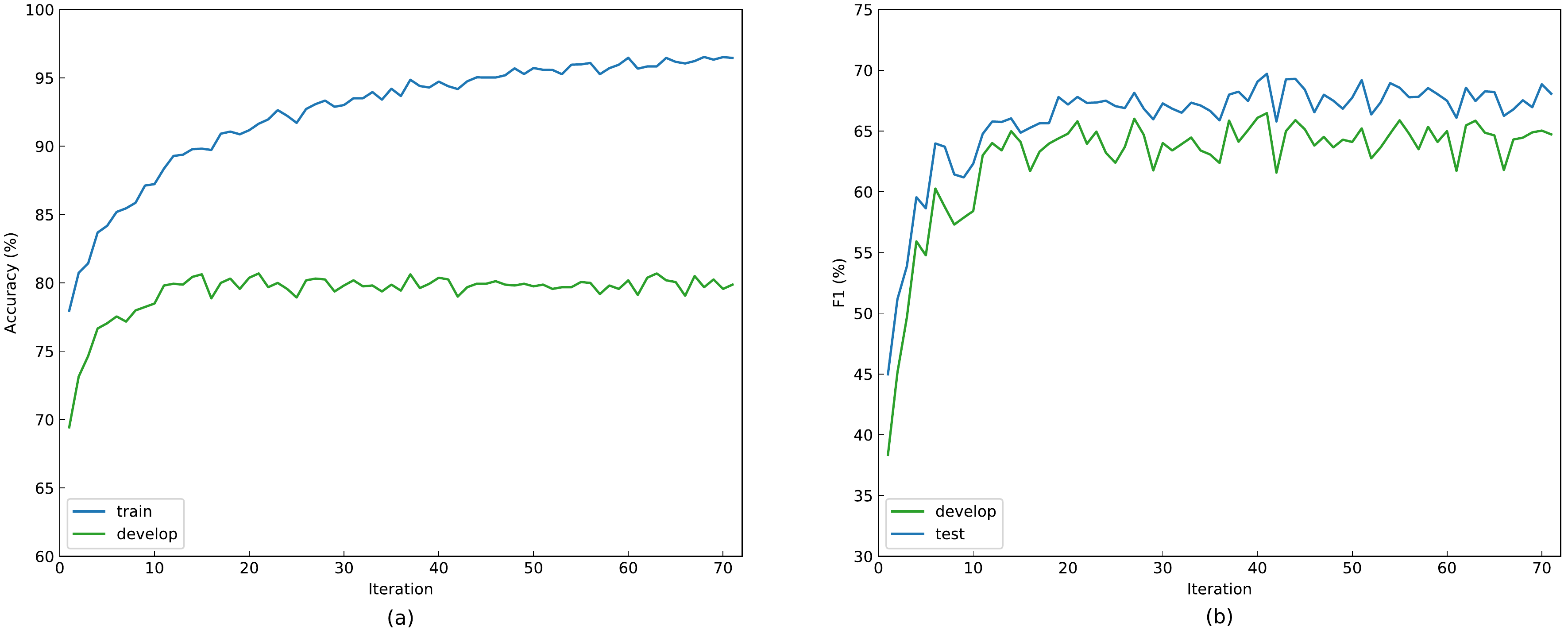}
    \caption{\label{fig:train_process}Training progress of the CNN-Multi model.}
\end{figure*}

\subsubsection{Comparison with baselines}

Previous methods \cite{DBLP:conf/coling/DSouzaN14,Sahu2016Relation,Raj2017} followed inconsistent data split schemes. To compare these to the proposed methods, all methods were reimplemented and evaluated using the official data split of the 2010 i2b2/VA relation corpus \cite{Uzuner2011}, as shown in Table~\ref{tab:relationcount}. All model hyperparameters remained unchanged during the reimplementation. To maintain a fair comparison, the part-of-speech and chunk features used in \cite{Sahu2016Relation} were removed, and word position embeddings were added to \citet{Raj2017}'s models. The performance results are displayed in Table~\ref{tab:classifiers}, including 95\% confidence intervals for each performance metric derived via bootstrapping \cite{DiCiccio1996}. The same bootstrapping method described in \cite{gao2017hierarchical} was used.

\paragraph{System performance} \citet{Rink2011} presented a support vector machine (SVM) method and achieved the best result in the 2010 i2b2/VA challenge. As the relation corpus available to the research community is a subset of that used during the 2010 i2b2/VA challenge, \citet{DBLP:conf/coling/DSouzaN14} re-implemented this method using the accessible dataset and obtains a F1 score of 62.1. They also proposed an improved single-model method and an ensemble-based method within an integer linear programming (ILP) framework, which became the new single-model and ensemble-based state-of-the-art methods, respectively. CNN achieves a slightly lower F1 score than SVM. CRNN-Max improved upon SVM, but still lags behind the single-model state-of-the-art method SVM+ILP. The models proposed in this paper outperformed SVM+ILP. CNN-Multi achieved a very similar result to Ensemble+ILP without depending on external features, and was significantly less complex to implement.

\begin{table}[!htb]
  \footnotesize
  \caption{\label{tab:classifiers}System performance comparison with other models using the 2010 i2b2/VA relation corpus.}
  \begin{threeparttable}
    \begin{tabularx}{\linewidth}{llXXX}
      \hline
      Classifier & \makecell[l]{External\\features} & P & R & F1 \\ \hline
      \multicolumn{5}{l}{\emph{Single-model methods}} \\
      SVM \cite{Rink2011} & Set1 & 66.7 & 58.1 & 62.1 \\
      SVM+ILP \cite{DBLP:conf/coling/DSouzaN14} & Set2 & 58.9 & \bf75.0 & 66.0 \\
      \makecell[l]{CNN \cite{Sahu2016Relation}\\$ $} & \makecell[l]{None\\$ $} & \makecell[l]{72.2\\(70.9, 73.5)} & \makecell[l]{54.1\\(52.8, 55.3)} & \makecell[l]{61.8\\(60.7, 62.9)} \\
      \makecell[l]{CRNN-Max \cite{Raj2017}\\$ $} & \makecell[l]{None\\$ $}& \makecell[l]{62.0\\(60.8, 63.1)} & \makecell[l]{64.6\\(63.4, 65.7)} & \makecell[l]{63.3\\(62.2, 64.3)} \\
      \makecell[l]{CRNN-Att \cite{Raj2017}\\$ $} & \makecell[l]{None\\$ $}& \makecell[l]{64.7\\(63.4, 65.9)} & \makecell[l]{56.5\\(55.3, 57.7)} & \makecell[l]{60.3\\(59.2, 61.4)} \\
      \makecell[l]{CNN-Max\\$ $}  & \makecell[l]{None\\$ $}& \makecell[l]{\bf73.4\\(72.2, 74.6)} & \makecell[l]{62.4\\(61.2, 63.6)} & \makecell[l]{67.5\\(66.4, 68.5)} \\
      \makecell[l]{CNN-Multi\\$ $}  & \makecell[l]{None\\$ $}& \makecell[l]{73.1\\(71.9, 74.2)} & \makecell[l]{66.7\\(65.4, 67.8)} & \makecell[l]{\bf69.7\\(68.7, 70.6)} \\
      \makecell[l]{CNN-Multi-C\\$ $}  & \makecell[l]{None\\$ $}& \makecell[l]{72.8\\(71.7, 74.0)} & \makecell[l]{65.9\\(64.7, 67.0)} & \makecell[l]{69.2\\(68.2, 70.2)} \\
      \hline
      \multicolumn{5}{l}{\emph{Ensemble-based method}} \\
      Ensemble+ILP \cite{DBLP:conf/coling/DSouzaN14} & set2 & 72.9 & 66.7 & 69.6 \\ \hline
    \end{tabularx}
    \begin{tablenotes}[para,flushleft]
      Set1: POS, chunk, semantic role labeler, word lemma, dependency parse, assertion type, sentiment category, Wikipedia; \\
      Set2: POS, chunk, semantic role labeler, word lemma, dependency parse, assertion type, sentiment category, Wikipedia, manually labeled patterns.
    \end{tablenotes}
  \end{threeparttable}
\end{table}

\paragraph{Category-wise performance} Table~\ref{tab:category} shows the performance of the neural network methods in the three relation categories. All methods demonstrated a better performance on \emph{medical problem-test} relations, potentially due to two conditions: (1) the relation type number of \emph{medical problem-treatment} relations is twice that of \emph{medical problem-test} relations; and (2) as shown in Table~\ref{tab:relationcount}, \emph{medical problem-medical problem} relations have a high relation type imbalance, which is adverse for classification. Compared with CNN-Max, CNN-Multi obtained significantly higher F1 scores for both \emph{medical problem-treatment} and \emph{medical problem-medical problem} relations, but improved much less for \emph{medical problem-test} relations. This may indicate the relative position information of features, extracted via multi-pooling operation, works well for relatively complex relation classifications whereas max-pooling is sufficient for simpler relation classifications. Among these neural network methods, CNN-Multi performed best for \emph{medical problem-treatment} and \emph{medical problem-medical problem} relations, whereas CNN-Multi-C performed best in \emph{medical problem-test} relations.

\begin{table}[!htb]
  \centering
  \footnotesize
  \caption{\label{tab:category}Category-wise performance comparison with other neural network models using the 2010 i2b2/VA relation corpus.}
  \begin{threeparttable}
    \begin{tabularx}{\linewidth}{lXXXXXXXXX}
      \hline Classifier & \multicolumn{3}{l}{TrP relations} & \multicolumn{3}{l}{TeP relations} & \multicolumn{3}{l}{PP relations} \\ \cmidrule(lr){2-4}\cmidrule(lr){5-7}\cmidrule(lr){8-10}
      & P & R & F1 & P & R & F1 & P & R & F1 \\ \hline
      CNN \cite{Sahu2016Relation} & 64.5 & 47.5 & 54.7 & 79.5 & 68.6 & 73.7 & 70.6 & 41.0 & 51.9 \\
      CRNN-Max \cite{Raj2017} & 53.9 & 60.2 & 56.9 & 68.7 & 77.1 & 72.7 & 65.3 & 51.4 & 57.5 \\
      CRNN-Att \cite{Raj2017} & 62.8 & 46.7 & 53.6 & 66.0 & 75.7 & 70.5 & 64.3 & 41.2 & 50.2 \\
      CNN-Max & 67.1 & 54.3 & 60.0 & 80.3 & 76.4 & 78.3 & 70.4 & 53.2 & 60.6 \\
      CNN-Multi & 68.1 & 60.0 & \bf63.8 & 77.9 & 79.3 & 78.6 & 72.0 & 56.9 & \bf63.6 \\
      CNN-Multi-C & 67.9 & 58.3 & 62.7 & 79.3 & 78.3 & \bf78.8 & 68.9 & 58.1 & 63.1 \\
      \hline
    \end{tabularx}
    \begin{tablenotes}[para,flushleft]
      TrP, Medical problem-Treatment; TeP, Medical problem-Test; PP, Medical problem-Medical problem.
    \end{tablenotes}
  \end{threeparttable}
\end{table}

\paragraph{Class-wise performance} Table~\ref{tab:subclasses} shows the performance of neural network methods for each positive relation type. In combination with Table~\ref{tab:relationcount}, this demonstrates that relation types with a small training size (\emph{TrIP}, 51; \emph{TrWP}, 24; \emph{TrNAP}, 62) provided poor performance, and class-wise performance improved as the training size increased.

\begin{sidewaystable}
  \centering
  \footnotesize
  \caption{\label{tab:subclasses} Class-wise performance comparison with other neural network models using the 2010 i2b2/VA relation corpus.}
  \begin{tabular*}{\textwidth}{lllllllllllllllllllllllll}
    \hline Classifier & \multicolumn{3}{l}{TrIP} & \multicolumn{3}{l}{TrWP} & \multicolumn{3}{l}{TrCP} & \multicolumn{3}{l}{TrAP} & \multicolumn{3}{l}{TrNAP} & \multicolumn{3}{l}{TeRP} & \multicolumn{3}{l}{TeCP} & \multicolumn{3}{l}{PIP} \\ \cmidrule(lr){2-4}\cmidrule(lr){5-7}\cmidrule(lr){8-10}\cmidrule(lr){11-13}\cmidrule(lr){14-16}\cmidrule(lr){17-19}\cmidrule(lr){20-22}\cmidrule(lr){23-25}
    & P & R & F1 & P & R & F1 & P & R & F1 & P & R & F1 & P & R & F1 & P & R & F1 & P & R & F1 & P & R & F1 \\ \hline
    CNN \cite{Sahu2016Relation} & 0.0 & 0.0 & 0.0 & 0.0 & 0.0 & 0.0 & 55.9 & 30.4 & 39.4 & 65.6 & 60.9 & 63.2 & 44.4 & 3.6 & 6.6 & 80.4 & 77.7 & 79.0 & 57.5 & 13.6 & 22.0 & 70.6 & 41.0 & 51.9 \\
    CRNN-Max \cite{Raj2017} & 35.8 & 12.5 & \bf18.5 & 0.0 & 0.0 & 0.0 & 43.9 & 41.8 & 42.8 & 55.7 & 75.6 & 64.1 & 100.0 & 0.9 & 1.8 & 74.4 & 83.0 & 78.5 & 35.6 & 41.4 & 38.3 & 65.3 & 51.4 & 57.5 \\
    CRNN-Att \cite{Raj2017} & 0.0 & 0.0 & 0.0 & 0.0 & 0.0 & 0.0 & 40.5 & 38.6 & 39.5 & 68.0 & 58.4 & 62.8 & 0.0 & 0.0 & 0.0 & 66.0 & 88.2 & 75.5 & 0.0 & 0.0 & 0.0 & 64.3 & 41.2 & 50.2 \\
    CNN-Max & 0.0 & 0.0 & 0.0 & 0.0 & 0.0 & 0.0 & 54.3 & 43.9 & \bf48.5 & 69.2 & 67.9 & 68.5 & 100.0 & 1.8 & 3.5 & 80.7 & 85.0 & 82.8 & 72.3 & 24.0 & 36.0 & 70.4 & 53.2 & 60.6 \\
    CNN-Multi & 50.0 & 2.0 & 3.8 & 0.0 & 0.0 & 0.0 & 56.7 & 41.8 & 48.1 & 69.6 & 76.2 & \bf72.8 & 100.0 & 2.7 & 5.2 & 77.8 & 88.9 & \bf83.0 & 81.6 & 21.0 & 33.4 & 72.0 & 56.9 & \bf63.6 \\
    CNN-Multi-C & 100.0 & 2.0 & 3.9 & 0.0 & 0.0 & 0.0 & 54.9 & 42.4 & 47.9 & 69.7 & 73.6 & 71.6 & 57.1 & 3.6 & \bf6.7 & 79.8 & 86.5 & \bf83.0 & 70.6 & 28.4 & \bf40.5 & 68.9 & 58.1 & 63.1 \\
    \hline
  \end{tabular*}
\end{sidewaystable}

\subsection{More analysis}

\paragraph{Model training progress} Fig.~\ref{fig:train_process}(a) shows accuracy curves of the CNN-Multi model. The accuracy curve of the training set maintained high values, indicating the model fit the dataset well. The accuracy of the development set generally tends to stabilize after 15 iterations. In Fig.~\ref{fig:train_process}(b), the F1 score curves of the development and test sets show similar trends; these curves appear less smooth because the training set was shuffled for each iteration. F1 score on the development set reached its optimal value at the 41st iteration, after which system parameters were maintained to evaluate system performance on the test set.

\paragraph{Errors} Table~\ref{tab:error} contains no cross-category misclassification, and relation samples are evidently more often misclassified as the relation type whose training size is larger. This is due to the fact that during multi-class model training, models have more offset for classes with larger training sizes. Considering this situation, sampling methods can be considered as a strategy to improve model performance in future works.

\begin{table*}[!htb]
  \centering
  \footnotesize
  \caption{\label{tab:error} Confusion matrix of the system output of the CNN-Multi model.}
  \begin{threeparttable}
    \begin{tabularx}{\linewidth}{lXXXXXXXXXXX}
      \hline & \multicolumn{11}{c}{System output} \\ \cmidrule{2-12}
       & TrIP & TrWP & TrCP & TrAP & TrNAP & NTrP & TeRP & TeCP & NTeP & PIP & NPP \\ \hline
      TrIP & \bf3 & & \emph{13} & \emph{93} & & \emph{43} & & & & & \\
      TrWP & & & \emph{11} & \emph{38} & & \emph{60} & & & & & \\
      TrCP & & & \bf143 & \emph{73} & & \emph{126} & & & & & \\
      TrAP & & & \emph{26} & \bf1320 & & \emph{386} & & & & & \\
      TrNAP & & & \emph{15} & \emph{53} & \bf3 & \emph{41} & & & & & \\
      NTrP & \emph{3} & & \emph{44} & \emph{319} & & \bf2393 & & & & & \\
      TeRP & & & & & & & \bf1831 & \emph{7} & \emph{222} & & \\
      TeCP & & & & & & & \emph{112} & \bf71 & \emph{155} & & \\
      NTeP & & & & & & & \emph{411} & \emph{9} & \bf1554 & & \\
      PIP & & & & & & & & & & \bf824 & \emph{624} \\
      NPP & & & & & & & & & & \emph{320} & \bf7769 \\
      \hline
    \end{tabularx}
    \begin{tablenotes}[para,flushleft]
      Zero items were removed from this table. Correctly classified items are bolded and the remainder are italicized.
    \end{tablenotes}
  \end{threeparttable}
\end{table*}

\paragraph{Effect of category-level constraint} As shown in Table~\ref{tab:classifiers}, system performance declined after adding a category-level constraint into the CNN-Multi model, whereas advantages of this constraint were not reflected. There are two potential reasons for this: (1) the CNN-Multi model was well trained and no cross-category misclassifications existed; and (2) the sample number of each relation type was too small for updating only the class vectors $W^{classes}_l$($l' \in \text{Category}_x$) during training to improve the generalization capability of the model. This constraint may be effective when experimenting with large datasets.

\begin{figure}[!htb]
    \centering
    \includegraphics[width=\linewidth]{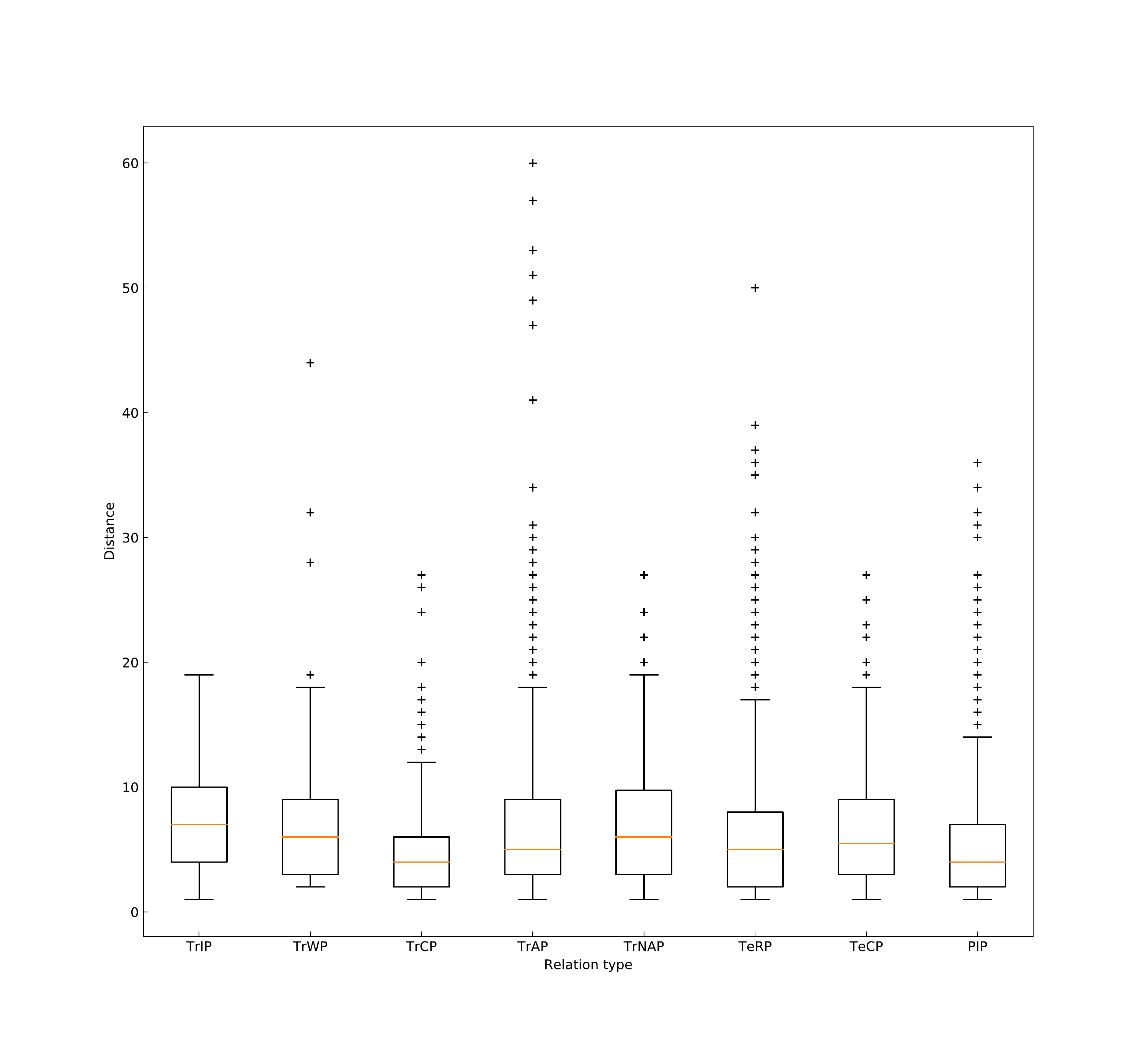}
    \caption{\label{fig:distance}Distance between medical concepts. Distance was calculated from the number of tokens between two medical concepts.}
\end{figure}

\paragraph{Distance between medical concepts} The distance between two medical concepts was calculated from the number of tokens between the concepts. Fig.~\ref{fig:distance} illustrates the distance distribution of different relation types. In most samples, the distance between concepts was less than 20 tokens, however, there are still some long-distance relations, which are more challenging to be classified.


\section{Related work}

Before deep learning research became popular, most relation classification tasks used statistical machine learning methods. Many researchers in the general and medical domains focused on feature-based and kernel-based methods \cite{Bunescu2005,Rink2010,Rink2011,Zhu2013,DBLP:conf/coling/DSouzaN14,Kim2015}, which are limited by conditions such as manual feature engineering and dependence on existing NLP toolkits.

More recently, researchers began investigating the performance of deep learning methods in relation classification tasks and achieved satisfactory results. Various deep architectures have been proposed for relation classification in the general domain, including the recurrent neural network (MV-RNN) \cite{Socher2012}, CNN with softmax classification \cite{Zeng2014}, factor-based compositional embedding model (FCM) \cite{yu2014factor}, and word embedding-based models \cite{hashimoto-EtAl:2015:CoNLL}. Many RNN- and CNN-based variants exist. Because the max-pooling operation in CNN models experiences significant linguistic feature losses in sentences, some researchers introduced dependency trees for this application such as bidirectional long short-term memory networks (BLSTM) \cite{Zhang2015a}, dependency-based neural networks (DepNN) \cite{liu-EtAl:2015:ACL-IJCNLP}, shortest dependency path-based CNN \cite{xu-EtAl:2015:EMNLP1}, long short term memory networks along shortest dependency paths (SDP-LSTM) \cite{Xu2015}, deep recurrent neural networks (DRNN) \cite{xu-EtAl:2016:COLING1}, and jointed sequential and tree-structured LSTM-RNN \cite{miwa-bansal:2016:P16-1}. Although the above studies achieved solid results, further research was devoted to eliminating the dependence on NLP parsers. \citet{DBLP:conf/acl/SantosXZ15} proposed a new pairwise ranking loss function where only two class representations were updated in every training round. Similarly, \citet{Wang2016} introduced a pairwise margin-based loss function and multi-level attention mechanism, achieving new state-of-the-art results for relation classification. Some feature-free neural network methods also exist for relation classification on biomedical and clinical text. \citet{Liu2016} employed CNN for drug-drug interaction (DDI) extraction, and \citet{Quan2016} proposed a multichannel convolutional neural network (MCCNN) for this task. Furthermore, several attention-based methods \cite{Asada2017,sahu2017drug,Zheng2017,Raj2017} were presented for automated biomedical relation extraction. This paper aimed is to train a feature-free CNN-based model for classifying medical relations in clinical records.

\section{Conclusion}

This paper presented a novel CNN-based model for classifying relations between medical concepts in clinical records. The model performed well classifying relations in the 2010 i2b2/VA relation corpus. A multi-pooling operation helped to extract more precise and richer features in the convolutional layer, indicating feature extraction based on concept pair positioning can improve the efficacy of relation classification on clinical records. Although this model was applied to relations between medical concepts in clinical records, it could also be adapted to classify relations in other domains.

\section*{Acknowledgments}
The authors are grateful to the Editors and the anonymous reviewers for their insightful comments. We would also like to thank the data support from the 2010 i2b2/VA challenge, as well as the useful discussions with Kehai Chen and Conghui Zhu.

\section*{References}
\bibliography{AIIM18HE}

\begin{thebibliography}{36}
\providecommand{\natexlab}[1]{#1}
\providecommand{\url}[1]{\texttt{#1}}
\providecommand{\href}[2]{#2}
\providecommand{\path}[1]{#1}
\providecommand{\eprint}[1]{\href{http://arxiv.org/abs/#1}{\path{#1}}}
\providecommand{\DOIprefix}{doi:}
\providecommand{\ArXivprefix}{arXiv:}
\providecommand{\URLprefix}{URL: }
\providecommand{\Pubmedprefix}{pmid:}
\providecommand{\doi}[1]{\href{http://dx.doi.org/#1}{\path{#1}}}
\providecommand{\Pubmed}[1]{\href{pmid:#1}{\path{#1}}}
\providecommand{\BIBand}{and}
\providecommand{\bibinfo}[2]{#2}
\ifx\xfnm\undefined \def\xfnm[#1]{\unskip,\space#1}\fi
\bibitem[{Hendrickx et~al.(2010)Hendrickx, Kim, Kozareva, Nakov, {{\'{O}}
  S{\'{e}}aghdha}, Pad{\'{o}} et~al.}]{Hendrickx2010}
\bibinfo{author}{Hendrickx\xfnm[ I.]}, \bibinfo{author}{Kim\xfnm[ S.N.]},
  \bibinfo{author}{Kozareva\xfnm[ Z.]}, \bibinfo{author}{Nakov\xfnm[ P.]},
  \bibinfo{author}{{{\'{O}} S{\'{e}}aghdha}\xfnm[ D.]},
  \bibinfo{author}{Pad{\'{o}}\xfnm[ S.]}, et~al.
\newblock \bibinfo{title}{{SemEval-2010 Task 8 : Multi-Way Classification of
  Semantic Relations Between Pairs of Nominals}}.
\newblock \bibinfo{journal}{Computational Linguistics}
  \bibinfo{year}{2010};(\bibinfo{number}{June 2009}):\bibinfo{pages}{94--99}.
\bibitem[{Uzuner et~al.(2011)Uzuner, South, Shen and DuVall}]{Uzuner2011}
\bibinfo{author}{Uzuner\xfnm[ {\"{O}}.]}, \bibinfo{author}{South\xfnm[ B.R.]},
  \bibinfo{author}{Shen\xfnm[ S.]}, \bibinfo{author}{DuVall\xfnm[ S.L.]}.
\newblock \bibinfo{title}{{2010 i2b2/VA challenge on concepts, assertions, and
  relations in clinical text}}.
\newblock \bibinfo{journal}{Journal of the American Medical Informatics
  Association}
  \bibinfo{year}{2011};\bibinfo{volume}{18}(\bibinfo{number}{5}):\bibinfo{pages}{552--556}.
\bibitem[{Socher et~al.(2012)Socher, Huval, Manning and Ng}]{Socher2012}
\bibinfo{author}{Socher\xfnm[ R.]}, \bibinfo{author}{Huval\xfnm[ B.]},
  \bibinfo{author}{Manning\xfnm[ C.D.]}, \bibinfo{author}{Ng\xfnm[ A.Y.]}.
\newblock \bibinfo{title}{{Semantic Compositionality through Recursive
  Matrix-Vector Spaces}}.
\newblock \bibinfo{journal}{Proceedings of the 2012 Joint Conference on
  Empirical Methods in Natural Language Processing and Computational Natural
  Language Learning}
  \bibinfo{year}{2012};(\bibinfo{number}{Mv}):\bibinfo{pages}{1201--1211}.
\bibitem[{Zeng et~al.(2014)Zeng, Liu, Lai, Zhou and Zhao}]{Zeng2014}
\bibinfo{author}{Zeng\xfnm[ D.]}, \bibinfo{author}{Liu\xfnm[ K.]},
  \bibinfo{author}{Lai\xfnm[ S.]}, \bibinfo{author}{Zhou\xfnm[ G.]},
  \bibinfo{author}{Zhao\xfnm[ J.]}.
\newblock \bibinfo{title}{{Relation Classification via Convolutional Deep
  Neural Network}}.
\newblock \bibinfo{journal}{COLING}
  \bibinfo{year}{2014};(\bibinfo{number}{2011}):\bibinfo{pages}{2335--2344}.
\bibitem[{dos Santos et~al.(2015)dos Santos, Xiang and
  Zhou}]{DBLP:conf/acl/SantosXZ15}
\bibinfo{author}{dos Santos\xfnm[ C.N.]}, \bibinfo{author}{Xiang\xfnm[ B.]},
  \bibinfo{author}{Zhou\xfnm[ B.]}.
\newblock \bibinfo{title}{{Classifying Relations by Ranking with Convolutional
  Neural Networks}}.
\newblock In: \bibinfo{booktitle}{Proceedings of the 53rd Annual Meeting of the
  Association for Computational Linguistics and the 7th International Joint
  Conference on Natural Language Processing (Volume 1: Long Papers)}.
  \bibinfo{number}{3}; \bibinfo{year}{2015}, p. \bibinfo{pages}{626--634}.
\bibitem[{Wang et~al.(2016)Wang, Cao, de~Melo and Liu}]{Wang2016}
\bibinfo{author}{Wang\xfnm[ L.]}, \bibinfo{author}{Cao\xfnm[ Z.]},
  \bibinfo{author}{de~Melo\xfnm[ G.]}, \bibinfo{author}{Liu\xfnm[ Z.]}.
\newblock \bibinfo{title}{{Relation Classification via Multi-Level Attention
  CNNs}}.
\newblock In: \bibinfo{booktitle}{Proceedings of the 54th Annual Meeting of the
  Association for Computational Linguistics (Volume 1: Long Papers)}.
  \bibinfo{year}{2016}, p. \bibinfo{pages}{1298--1307}.
\bibitem[{Liu et~al.(2016)Liu, Tang, Chen and Wang}]{Liu2016}
\bibinfo{author}{Liu\xfnm[ S.]}, \bibinfo{author}{Tang\xfnm[ B.]},
  \bibinfo{author}{Chen\xfnm[ Q.]}, \bibinfo{author}{Wang\xfnm[ X.]}.
\newblock \bibinfo{title}{{Drug-Drug Interaction Extraction via Convolutional
  Neural Networks}}.
\newblock \bibinfo{journal}{Computational and Mathematical Methods in Medicine}
  \bibinfo{year}{2016};\bibinfo{volume}{2016}.
\bibitem[{Zhao et~al.(2016)Zhao, Yang, Luo, Lin and Wang}]{Zhao2016}
\bibinfo{author}{Zhao\xfnm[ Z.]}, \bibinfo{author}{Yang\xfnm[ Z.]},
  \bibinfo{author}{Luo\xfnm[ L.]}, \bibinfo{author}{Lin\xfnm[ H.]},
  \bibinfo{author}{Wang\xfnm[ J.]}.
\newblock \bibinfo{title}{{Drug drug interaction extraction from biomedical
  literature using syntax convolutional neural network}}.
\newblock \bibinfo{journal}{Bioinformatics}
  \bibinfo{year}{2016};\bibinfo{volume}{32}(\bibinfo{number}{22}):\bibinfo{pages}{3444--3453}.
\bibitem[{Quan et~al.(2016)Quan, Hua, Sun and Bai}]{Quan2016}
\bibinfo{author}{Quan\xfnm[ C.]}, \bibinfo{author}{Hua\xfnm[ L.]},
  \bibinfo{author}{Sun\xfnm[ X.]}, \bibinfo{author}{Bai\xfnm[ W.]}.
\newblock \bibinfo{title}{{Multichannel convolutional neural network for
  biological relation extraction}}.
\newblock \bibinfo{journal}{BioMed Research International}
  \bibinfo{year}{2016};\bibinfo{volume}{2016}.
\bibitem[{Asada et~al.(2017)Asada, Miwa and Sasaki}]{Asada2017}
\bibinfo{author}{Asada\xfnm[ M.]}, \bibinfo{author}{Miwa\xfnm[ M.]},
  \bibinfo{author}{Sasaki\xfnm[ Y.]}.
\newblock \bibinfo{title}{{Extracting Drug-Drug Interactions with Attention
  CNNs}}.
\newblock In: \bibinfo{booktitle}{BioNLP 2017}. \bibinfo{year}{2017}, p.
  \bibinfo{pages}{9--18}.
\bibitem[{Sahu and Anand(2017)}]{sahu2017drug}
\bibinfo{author}{Sahu\xfnm[ S.K.]}, \bibinfo{author}{Anand\xfnm[ A.]}.
\newblock \bibinfo{title}{{Drug-Drug Interaction Extraction from Biomedical
  Text Using Long Short Term Memory Network}}.
\newblock \bibinfo{journal}{arXiv preprint arXiv:170108303}
  \bibinfo{year}{2017};.
\bibitem[{Zheng et~al.(2017)Zheng, Lin, Luo, Zhao, Li, Zhang
  et~al.}]{Zheng2017}
\bibinfo{author}{Zheng\xfnm[ W.]}, \bibinfo{author}{Lin\xfnm[ H.]},
  \bibinfo{author}{Luo\xfnm[ L.]}, \bibinfo{author}{Zhao\xfnm[ Z.]},
  \bibinfo{author}{Li\xfnm[ Z.]}, \bibinfo{author}{Zhang\xfnm[ Y.]}, et~al.
\newblock \bibinfo{title}{{An attention-based effective neural model for
  drug-drug interactions extraction}}.
\newblock \bibinfo{journal}{BMC Bioinformatics}
  \bibinfo{year}{2017};\bibinfo{volume}{18}(\bibinfo{number}{1}).
\bibitem[{Sahu et~al.(2016)Sahu, Anand, Oruganty and Gattu}]{Sahu2016Relation}
\bibinfo{author}{Sahu\xfnm[ S.K.]}, \bibinfo{author}{Anand\xfnm[ A.]},
  \bibinfo{author}{Oruganty\xfnm[ K.]}, \bibinfo{author}{Gattu\xfnm[ M.]}.
\newblock \bibinfo{title}{{Relation extraction from clinical texts using domain
  invariant convolutional neural network}}.
\newblock \bibinfo{journal}{Proceedings of the 15th Workshop on Biomedical
  Natural Language Processing} \bibinfo{year}{2016};:\bibinfo{pages}{71}.
\bibitem[{Raj et~al.(2017)Raj, Sahu and Anand}]{Raj2017}
\bibinfo{author}{Raj\xfnm[ D.]}, \bibinfo{author}{Sahu\xfnm[ S.K.]},
  \bibinfo{author}{Anand\xfnm[ A.]}.
\newblock \bibinfo{title}{{Learning local and global contexts using a
  convolutional recurrent network model for relation classification in
  biomedical text}}.
\newblock \bibinfo{journal}{Proceedings of the 21st Conference on Computational
  Natural Language Learning (CoNLL 2017)}
  \bibinfo{year}{2017};:\bibinfo{pages}{311--321}.
\bibitem[{Chen et~al.(2015)Chen, Xu, Liu, Zeng and Zhao}]{ChenXLZ015}
\bibinfo{author}{Chen\xfnm[ Y.]}, \bibinfo{author}{Xu\xfnm[ L.]},
  \bibinfo{author}{Liu\xfnm[ K.]}, \bibinfo{author}{Zeng\xfnm[ D.]},
  \bibinfo{author}{Zhao\xfnm[ J.]}.
\newblock \bibinfo{title}{{Event Extraction via Dynamic Multi-Pooling
  Convolutional Neural Networks}}.
\newblock In: \bibinfo{booktitle}{Proceedings of the 53rd Annual Meeting of the
  Association for Computational Linguistics and the 7th International Joint
  Conference on Natural Language Processing}. \bibinfo{year}{2015}, p.
  \bibinfo{pages}{167--176}.
\bibitem[{Zhang et~al.(2015{\natexlab{a}})Zhang, Zhang and Hao}]{Zhang2015}
\bibinfo{author}{Zhang\xfnm[ J.]}, \bibinfo{author}{Zhang\xfnm[ D.]},
  \bibinfo{author}{Hao\xfnm[ J.]}.
\newblock \bibinfo{title}{{Local translation prediction with global sentence
  representation}}.
\newblock In: \bibinfo{booktitle}{International Joint Conference on Artificial
  Intelligence}; vol. \bibinfo{volume}{2015-Janua}.
  \bibinfo{year}{2015}{\natexlab{a}}, p. \bibinfo{pages}{1398--1404}.
\bibitem[{Mikolov et~al.(2013)Mikolov, Corrado, Chen and Dean}]{Mikolov2013}
\bibinfo{author}{Mikolov\xfnm[ T.]}, \bibinfo{author}{Corrado\xfnm[ G.]},
  \bibinfo{author}{Chen\xfnm[ K.]}, \bibinfo{author}{Dean\xfnm[ J.]}.
\newblock \bibinfo{title}{{Efficient Estimation of Word Representations in
  Vector Space}}.
\newblock \bibinfo{journal}{Proceedings of the International Conference on
  Learning Representations} \bibinfo{year}{2013};:\bibinfo{pages}{1--12}.
\bibitem[{Johnson et~al.(2016)Johnson, Pollard, Shen, Lehman, Feng, Ghassemi
  et~al.}]{Johnson2016}
\bibinfo{author}{Johnson\xfnm[ A.E.]}, \bibinfo{author}{Pollard\xfnm[ T.J.]},
  \bibinfo{author}{Shen\xfnm[ L.]}, \bibinfo{author}{Lehman\xfnm[ L.W.H.]},
  \bibinfo{author}{Feng\xfnm[ M.]}, \bibinfo{author}{Ghassemi\xfnm[ M.]},
  et~al.
\newblock \bibinfo{title}{{MIMIC-III, a freely accessible critical care
  database}}.
\newblock \bibinfo{journal}{Scientific Data}
  \bibinfo{year}{2016};\bibinfo{volume}{3}.
\bibitem[{Glorot and Bengio(2010)}]{DBLP:journals/jmlr/GlorotB10}
\bibinfo{author}{Glorot\xfnm[ X.]}, \bibinfo{author}{Bengio\xfnm[ Y.]}.
\newblock \bibinfo{title}{{Understanding the difficulty of training deep
  feedforward neural networks}}.
\newblock In: \bibinfo{booktitle}{Proceedings of the 13th International
  Conference on Artificial Intelligence and Statistics (AISTATS)};
  vol.~\bibinfo{volume}{9}. \bibinfo{year}{2010}, p. \bibinfo{pages}{249--256}.
\bibitem[{Srivastava et~al.(2014)Srivastava, Hinton, Krizhevsky, Sutskever and
  Salakhutdinov}]{Srivastava2014}
\bibinfo{author}{Srivastava\xfnm[ N.]}, \bibinfo{author}{Hinton\xfnm[ G.]},
  \bibinfo{author}{Krizhevsky\xfnm[ A.]}, \bibinfo{author}{Sutskever\xfnm[
  I.]}, \bibinfo{author}{Salakhutdinov\xfnm[ R.]}.
\newblock \bibinfo{title}{{Dropout: A Simple Way to Prevent Neural Networks
  from Overfitting}}.
\newblock \bibinfo{journal}{Journal of Machine Learning Research}
  \bibinfo{year}{2014};\bibinfo{volume}{15}:\bibinfo{pages}{1929--1958}.
\bibitem[{Souza and Ng(2014)}]{DBLP:conf/coling/DSouzaN14}
\bibinfo{author}{Souza\xfnm[ J.D.]}, \bibinfo{author}{Ng\xfnm[ V.]}.
\newblock \bibinfo{title}{{Ensemble-Based Medical Relation Classification}}.
\newblock In: \bibinfo{editor}{Hajic\xfnm[ J.]}, \bibinfo{editor}{Tsujii\xfnm[
  J.]}, editors. \bibinfo{booktitle}{COLING}. \bibinfo{publisher}{ACL};
  \bibinfo{year}{2014}, p. \bibinfo{pages}{1682--1693}.
\bibitem[{DiCiccio and Efron(1996)}]{DiCiccio1996}
\bibinfo{author}{DiCiccio\xfnm[ T.J.]}, \bibinfo{author}{Efron\xfnm[ B.]}.
\newblock \bibinfo{title}{{Bootstrap confidence intervals}}.
\newblock \bibinfo{journal}{Statistical Science}
  \bibinfo{year}{1996};\bibinfo{volume}{11}(\bibinfo{number}{3}):\bibinfo{pages}{189--228}.
\bibitem[{Gao et~al.(2017)Gao, Young, Qiu, Yoon, Christian, Fearn
  et~al.}]{gao2017hierarchical}
\bibinfo{author}{Gao\xfnm[ S.]}, \bibinfo{author}{Young\xfnm[ M.T.]},
  \bibinfo{author}{Qiu\xfnm[ J.X.]}, \bibinfo{author}{Yoon\xfnm[ H.J.]},
  \bibinfo{author}{Christian\xfnm[ J.B.]}, \bibinfo{author}{Fearn\xfnm[ P.A.]},
  et~al.
\newblock \bibinfo{title}{{Hierarchical attention networks for information
  extraction from cancer pathology reports}}.
\newblock \bibinfo{journal}{Journal of the American Medical Informatics
  Association} \bibinfo{year}{2017};.
\bibitem[{Rink et~al.(2011)Rink, Harabagiu and Roberts}]{Rink2011}
\bibinfo{author}{Rink\xfnm[ B.]}, \bibinfo{author}{Harabagiu\xfnm[ S.]},
  \bibinfo{author}{Roberts\xfnm[ K.]}.
\newblock \bibinfo{title}{{Automatic extraction of relations between medical
  concepts in clinical texts}}.
\newblock \bibinfo{journal}{Journal of the American Medical Informatics
  Association}
  \bibinfo{year}{2011};\bibinfo{volume}{18}(\bibinfo{number}{5}):\bibinfo{pages}{594--600}.
\bibitem[{Bunescu and Mooney(2005)}]{Bunescu2005}
\bibinfo{author}{Bunescu\xfnm[ R.]}, \bibinfo{author}{Mooney\xfnm[ R.]}.
\newblock \bibinfo{title}{{A shortest path dependency kernel for relation
  extraction}}.
\newblock In: \bibinfo{booktitle}{Human Language Technology Conference and
  Conference on Empirical Methods in Natural Language Processing}.
  \bibinfo{year}{2005}, p. \bibinfo{pages}{724--731}.
\bibitem[{Rink and Harabagiu(2010)}]{Rink2010}
\bibinfo{author}{Rink\xfnm[ B.]}, \bibinfo{author}{Harabagiu\xfnm[ S.]}.
\newblock \bibinfo{title}{{UTD: Classifying Semantic Relations by Combining
  Lexical and Semantic Resources}}.
\newblock \bibinfo{journal}{Proceedings of the 5th International Workshop on
  Semantic Evaluation}
  \bibinfo{year}{2010};(\bibinfo{number}{July}):\bibinfo{pages}{256--259}.
\bibitem[{Zhu et~al.(2013)Zhu, Cherry, Kiritchenko, Martin and
  de~Bruijn}]{Zhu2013}
\bibinfo{author}{Zhu\xfnm[ X.]}, \bibinfo{author}{Cherry\xfnm[ C.]},
  \bibinfo{author}{Kiritchenko\xfnm[ S.]}, \bibinfo{author}{Martin\xfnm[ J.]},
  \bibinfo{author}{de~Bruijn\xfnm[ B.]}.
\newblock \bibinfo{title}{{Detecting concept relations in clinical text:
  Insights from a state-of-the-art model}}.
\newblock \bibinfo{journal}{Journal of Biomedical Informatics}
  \bibinfo{year}{2013};\bibinfo{volume}{46}(\bibinfo{number}{2}):\bibinfo{pages}{275--285}.
\bibitem[{Kim et~al.(2015)Kim, Choe and Mueller}]{Kim2015}
\bibinfo{author}{Kim\xfnm[ J.]}, \bibinfo{author}{Choe\xfnm[ Y.]},
  \bibinfo{author}{Mueller\xfnm[ K.]}.
\newblock \bibinfo{title}{{Extracting clinical relations in electronic health
  records using enriched parse trees}}.
\newblock In: \bibinfo{booktitle}{Procedia Computer Science};
  vol.~\bibinfo{volume}{53}. \bibinfo{year}{2015}, p.
  \bibinfo{pages}{274--283}.
\bibitem[{Yu et~al.(2014)Yu, Gormley and Dredze}]{yu2014factor}
\bibinfo{author}{Yu\xfnm[ M.]}, \bibinfo{author}{Gormley\xfnm[ M.R.]},
  \bibinfo{author}{Dredze\xfnm[ M.]}.
\newblock \bibinfo{title}{{Factor-based Compositional Embedding Models}}.
\newblock \bibinfo{journal}{NIPS Workshop on Learning Semantics}
  \bibinfo{year}{2014};:\bibinfo{pages}{95--101}.
\bibitem[{Hashimoto et~al.(2015)Hashimoto, Stenetorp, Miwa and
  Tsuruoka}]{hashimoto-EtAl:2015:CoNLL}
\bibinfo{author}{Hashimoto\xfnm[ K.]}, \bibinfo{author}{Stenetorp\xfnm[ P.]},
  \bibinfo{author}{Miwa\xfnm[ M.]}, \bibinfo{author}{Tsuruoka\xfnm[ Y.]}.
\newblock \bibinfo{title}{{Task-Oriented Learning of Word Embeddings for
  Semantic Relation Classification}}.
\newblock In: \bibinfo{booktitle}{Proceedings of the Nineteenth Conference on
  Computational Natural Language Learning}. \bibinfo{address}{Beijing, China}:
  \bibinfo{publisher}{Association for Computational Linguistics};
  \bibinfo{year}{2015}, p. \bibinfo{pages}{268--278}.
\bibitem[{Zhang et~al.(2015{\natexlab{b}})Zhang, Zheng, Hu and
  Yang}]{Zhang2015a}
\bibinfo{author}{Zhang\xfnm[ S.]}, \bibinfo{author}{Zheng\xfnm[ D.]},
  \bibinfo{author}{Hu\xfnm[ X.]}, \bibinfo{author}{Yang\xfnm[ M.]}.
\newblock \bibinfo{title}{{Bidirectional Long Short-Term Memory Networks for
  Relation Classification}}.
\newblock In: \bibinfo{booktitle}{PACLIC}. \bibinfo{year}{2015}{\natexlab{b}},
  p. \bibinfo{pages}{73--78}.
\bibitem[{Liu et~al.(2015)Liu, Wei, Li, Ji, Zhou and
  WANG}]{liu-EtAl:2015:ACL-IJCNLP}
\bibinfo{author}{Liu\xfnm[ Y.]}, \bibinfo{author}{Wei\xfnm[ F.]},
  \bibinfo{author}{Li\xfnm[ S.]}, \bibinfo{author}{Ji\xfnm[ H.]},
  \bibinfo{author}{Zhou\xfnm[ M.]}, \bibinfo{author}{WANG\xfnm[ H.]}.
\newblock \bibinfo{title}{{A Dependency-Based Neural Network for Relation
  Classification}}.
\newblock In: \bibinfo{booktitle}{Proceedings of the 53rd Annual Meeting of the
  Association for Computational Linguistics and the 7th International Joint
  Conference on Natural Language Processing (Volume 2: Short Papers)}.
  \bibinfo{address}{Beijing, China}: \bibinfo{publisher}{Association for
  Computational Linguistics}; \bibinfo{year}{2015}, p.
  \bibinfo{pages}{285--290}.
\bibitem[{Xu et~al.(2015{\natexlab{a}})Xu, Feng, Huang and
  Zhao}]{xu-EtAl:2015:EMNLP1}
\bibinfo{author}{Xu\xfnm[ K.]}, \bibinfo{author}{Feng\xfnm[ Y.]},
  \bibinfo{author}{Huang\xfnm[ S.]}, \bibinfo{author}{Zhao\xfnm[ D.]}.
\newblock \bibinfo{title}{{Semantic Relation Classification via Convolutional
  Neural Networks with Simple Negative Sampling}}.
\newblock In: \bibinfo{booktitle}{Proceedings of the 2015 Conference on
  Empirical Methods in Natural Language Processing}. \bibinfo{address}{Lisbon,
  Portugal}: \bibinfo{publisher}{Association for Computational Linguistics};
  \bibinfo{year}{2015}{\natexlab{a}}, p. \bibinfo{pages}{536--540}.
\bibitem[{Xu et~al.(2015{\natexlab{b}})Xu, Mou, Li, Chen, Peng and
  Jin}]{Xu2015}
\bibinfo{author}{Xu\xfnm[ Y.]}, \bibinfo{author}{Mou\xfnm[ L.]},
  \bibinfo{author}{Li\xfnm[ G.]}, \bibinfo{author}{Chen\xfnm[ Y.]},
  \bibinfo{author}{Peng\xfnm[ H.]}, \bibinfo{author}{Jin\xfnm[ Z.]}.
\newblock \bibinfo{title}{{Classifying Relations via Long Short Term Memory
  Networks along Shortest Dependency Paths}}.
\newblock In: \bibinfo{booktitle}{Proceedings of the 2015 Conference on
  Empirical Methods in Natural Language Processing}.
  \bibinfo{year}{2015}{\natexlab{b}}, p. \bibinfo{pages}{1785--1794}.
\bibitem[{Xu et~al.(2016)Xu, Jia, Mou, Li, Chen, Lu
  et~al.}]{xu-EtAl:2016:COLING1}
\bibinfo{author}{Xu\xfnm[ Y.]}, \bibinfo{author}{Jia\xfnm[ R.]},
  \bibinfo{author}{Mou\xfnm[ L.]}, \bibinfo{author}{Li\xfnm[ G.]},
  \bibinfo{author}{Chen\xfnm[ Y.]}, \bibinfo{author}{Lu\xfnm[ Y.]}, et~al.
\newblock \bibinfo{title}{{Improved relation classification by deep recurrent
  neural networks with data augmentation}}.
\newblock In: \bibinfo{booktitle}{Proceedings of COLING 2016, the 26th
  International Conference on Computational Linguistics: Technical Papers}.
  \bibinfo{address}{Osaka, Japan}: \bibinfo{publisher}{The COLING 2016
  Organizing Committee}; \bibinfo{year}{2016}, p. \bibinfo{pages}{1461--1470}.
\bibitem[{Miwa and Bansal(2016)}]{miwa-bansal:2016:P16-1}
\bibinfo{author}{Miwa\xfnm[ M.]}, \bibinfo{author}{Bansal\xfnm[ M.]}.
\newblock \bibinfo{title}{{End-to-End Relation Extraction using LSTMs on
  Sequences and Tree Structures}}.
\newblock In: \bibinfo{booktitle}{Proceedings of the 54th Annual Meeting of the
  Association for Computational Linguistics (Volume 1: Long Papers)}.
  \bibinfo{address}{Berlin, Germany}: \bibinfo{publisher}{Association for
  Computational Linguistics}; \bibinfo{year}{2016}, p.
  \bibinfo{pages}{1105--1116}.

\end{thebibliography}
\bibliographystyle{AIIM}

\end{document}